\documentclass[letterpaper, 10 pt, conference]{ieeeconf}  

\IEEEoverridecommandlockouts                              

\usepackage{amsmath} 
\usepackage{xcolor}
\usepackage{svg}
\usepackage{graphicx}
\usepackage{caption}
\usepackage{subcaption}

\title{\LARGE \bf Multi-Camera Asynchronous Ball Localization and Trajectory Prediction with Factor Graphs and Human Poses}

\author{Qingyu Xiao$^{1}$, Zulfiqar Zaidi$^1$ and Matthew Gombolay$^{1}$
\thanks{$^{1}$ {All authors are affiliated with Georgia Institute of Technology}}
}

\begin{document}

\maketitle
\thispagestyle{empty}
\pagestyle{empty}

\begin{abstract}    
The rapid and precise localization and prediction of a ball are critical for developing agile robots in ball sports, particularly in sports like tennis characterized by high-speed ball movements and powerful spins. The Magnus effect induced by spin adds complexity to trajectory prediction during flight and bounce dynamics upon contact with the ground. In this study, we introduce an innovative approach that combines a multi-camera system with factor graphs for real-time and asynchronous 3D tennis ball localization. Additionally, we estimate hidden states like velocity and spin for trajectory prediction. Furthermore, to enhance spin inference early in the ball's flight, where limited observations are available, we integrate human pose data using a temporal convolutional network (TCN) to compute spin priors within the factor graph. This refinement provides more accurate spin priors at the beginning of the factor graph, leading to improved early-stage hidden state inference for prediction. Our result shows the trained TCN can predict the spin priors with RMSE of 5.27 Hz. Integrating TCN into the factor graph reduces the prediction error of landing positions by over 63.6\% compared to a baseline method that utilized an adaptive extended Kalman filter.


\end{abstract}

\section{INTRODUCTION}
Object tracking is a well-studied problem in the field of computer vision \cite{objecttracking_survey}. However, when it comes to tracking small and fast-moving objects, we find ourselves in a relatively unexplored territory. This presents a unique set of challenges, such as the limited time available for inference and the occurrence of motion blur due to the high speeds involved \cite{smallandfastmoving_objecttracking}. Development of algorithms capable of effectively tracking and predicting the trajectories of these fast-moving objects holds tremendous potential across various domains including robots throwing objects in factories~\cite{frank2012automated}, dynamic obstacle avoidance for autonomous driving~\cite{luo2018porca}, drone tracking~\cite{guvenc2018detection}, as well as estimating the trajectory of balls in sports, such as tennis, table tennis~\cite{gomez2020real,d2023robotic,abeyruwan2023sim2real}, soccer~\cite{nurrohmah2020detecting}, cricket \cite{cricket_traj}, baseball \cite{baseball_traj}, and many others \cite{kensrud2010determining}.

Tennis, as an example, is such a sport where precise ball trajectory estimation and prediction are of extreme importance \cite{tennis_timepressure}. Human players develop these skills through years of dedicated training. However, for an autonomous robotic system to excel in tennis, it must accurately anticipate the trajectory of an incoming ball and swiftly position itself. Previous studies \cite{zaidi2023athletic,yang2022varsm} have demonstrated that an autonomous robotic system can return a ball effectively when faced with non-aggressive and cooperative human shots. Nevertheless, when challenged by highly skilled players capable of delivering high-velocity shots with ``heavy spin,'' the vision system encounters difficulties in precisely determining the current and future ball positions with sufficient lead time for effective robot motion planning. 



    \begin{figure}[t]
      \centering
      \includegraphics[width=0.9\linewidth]{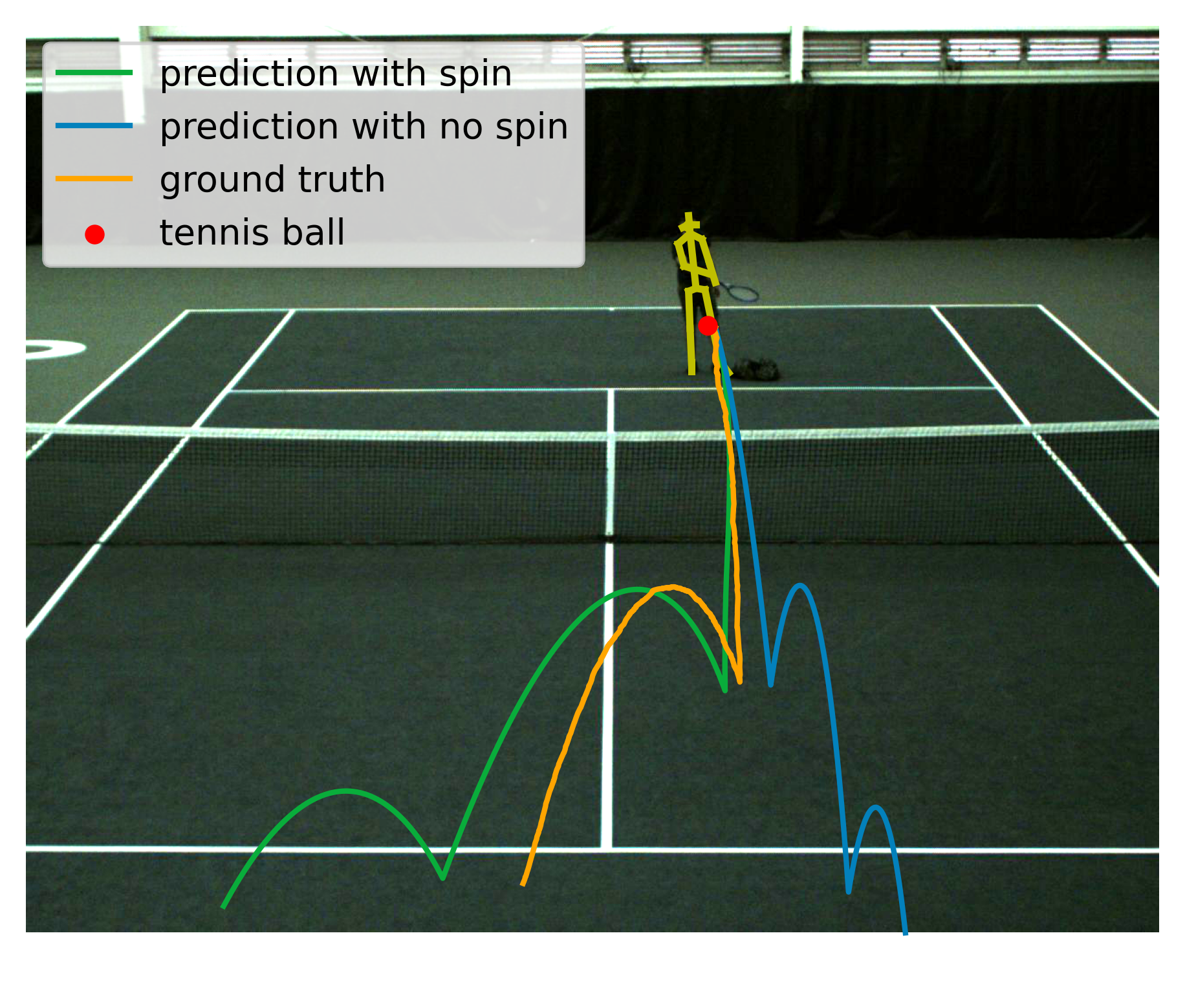}
      \vspace{-5mm}
      \caption{Leveraging time-series human pose data for ball spin estimation and fusing asynchronous camera detections through factor graphs substantially improves performance of trajectory prediction.}
      \label{fig: approach_overview}
    \end{figure}


Prior work has employed a range of techniques for tracking balls~\cite{ren2009tracking,wong2023multi,proximate_hrc}. However, these methods often rely on precise synchronization between paired cameras to triangulate the ball's location, which is difficult to achieve. Attempts to compensate using time filters can result in either data loss or increased localization errors. Stereo cameras \cite{zaidi2023athletic,yang2022varsm} provide more precise synchronization but face challenges when the ball is at a distance, leading to localization errors. Some work has used a single camera and leverage the known ball diameter for distance estimation~\cite{van20223d}. However, this method has practical limitations, especially with smaller balls at distance, such as in tennis.


Our motivation of this paper is to address the challenge of multi-camera asynchronous ball localization while simultaneously enhancing the accuracy of trajectory predictions. 
This paper explores 3D tennis ball localization and prediction by employing a factor graph framework, with human poses to compute spin priors. 
Factor graphs have traditionally found their niche in the field of robotics, particularly in the context of simultaneous localization and mapping (SLAM). SLAM involves robots fusing data from IMU sensors, capturing their dynamic motions, and stationary landscape landmarks from cameras to estimate their positions \cite{dellaert2017factor}. Drawing a parallel, we conceptualize the tennis ball as a dynamic landmark influenced by aerodynamic forces and restitutional bounces. We associate stationary camera detections with timestamps and connect them through dynamic factors derived from a physics-based dynamics model. This approach allows us to simultaneously estimate hidden states such as the ball's velocity and spin, significantly refining prediction accuracy. To achieve precise early-stage hidden state estimation in the ball's trajectory, we incorporate a temporal convolutional network (TCN) trained on human poses. In our final implementation, we employ GTSAM \cite{dellaert2012factor} in conjunction with the efficient incremental solver ISAM \cite{kaess2008isam,kaess2012isam2} to handle the inherent nonlinearity of the factor graph model.
The principal contributions of our work are summarized below:
\begin{itemize}
    \item Developed an asynchronous approach utilizing an incremental factor graph achieving fast and accurate estimation of the tennis ball's location, velocity and spin.
    \item Derived potential functions of the factors based on the existing research findings and incorporated into our proposed factor graph.
    \item Incorporated human poses to calculate spin priors for the factor graph, facilitating an early and precise estimation of the ball's trajectory. The integration of spin priors led to a substantial 63.6\% reduction in the RMSE for the ball's landing position vs.~baseline methods.
\end{itemize}

\section{Related Work}

\subsection{Ball Localization}


Various ball localization techniques have been reported in literature, typically using multi-camera systems and use triangulation to estimate the ball's location~\cite{ren2009tracking,fazio2018tennis,tebbe2019table, reatimeballtracking_gomez2019}. These approaches distributecameras around to maximize coverage of the target area but relies on synchronized image capture between paired cameras, which can lead to errors if the tracked object is moving rapidly. In contrast, our approach does not require synchronized frames from the paired cameras, avoiding this issue.


Other work has used stereo cameras for estimating the 3D coordinates of the ball \cite{zaidi2023athletic,yang2022varsm}. This method minimizes synchronization discrepancies, as the cameras are interconnected at a hardware level. However, when computing the locations of an object situated at considerable distances from the cameras—several meters or more—this technique encounters noise issues. The proximity of stereo cameras to each other (the small distance between two cameras) results in very small disparities between the captured images. Consequently, even minor noise at a one- or two-pixel level becomes significant compared to the small disparity, leading to substantial errors in the localization process~\cite{Olson2009}. In contrast, our approach places cameras at a significant distance from each other, which alleviates this issue.

Another approach \cite{van20223d} utilizes a single calibrated camera for ball localization in basketball. It detects the ball's centroid and apparent diameter within the image, using the latter as a metric to approximate the ball-camera distance. While effective for larger balls and close-range scenarios in sports like soccer or basketball, this method encounters difficulties with smaller spherical objects like tennis balls. The reduced ball size complicates accurate diameter estimation, making it less reliable for distance approximation. Instaed, our approach avoids this limitation by not relying on estimating ball size.


\subsection{Ball Dynamics and Prediction}

Accurate ball trajectory prediction hinges on two critical factors: aerodynamics during flight and bounce dynamics.
Aerodynamics during ball flight have been well-studied, considering effects such as drag, Magnus, and lifting forces~\cite{cooke2000overview, cross2014measurements}. Lift force, in particular, causes the ball's path to curve, adding complexity to prediction. Ball bounce behavior is equally vital. Inbound velocity and spin at the moment of contact profoundly impact the ball's subsequent path, leading to various outcomes. Prior studies \cite{cross2002measurements} have formulated equations for inbound and outbound dynamics, aiding prediction. While prior robotics-focused works use simpler bounce models \cite{zaidi2023athletic,tebbe2019table,tebbe2019table}, they struggle to predict spin-heavy ball trajectories post-bounce. Our approach accounts for these complexities, enhancing ball trajectory prediction.




\subsection{Stroke and Spin detection}
Various methods have been documented that utilize ball trajectory  to detect the strokes of table tennis players \cite{kulkarni2023table}, as well as to infer the spin on the ball \cite{tebbe2020spin}. An alternative technique proposed in \cite{gossard2023spindoe} involves applying non-regulation markings to the ball and then tracking these markings to precisely estimate the ball's spin. Nevertheless, this method's use of non-regulation markings renders it incompatible with standard equipment. Additionally, previous research has established a significant causal relationship between the executed stroke and the resulting spin on the ball \cite{spin_tennis_stroke}. Therefore, techniques employed for stroke detection can be repurposed to estimate spin dynamics as well. Several studies \cite{blank2017ball,zhao2019tenniseye,blank2015sensor} have attempted to detect human strokes using IMU sensors mounted directly onto the racket. However, this approach might be perceived as cumbersome and aesthetically inattractive, potentially hindering its wide acceptance and usage. In a more recent study , innovative method have emerged that leverage the analysis of human pose sequences to identify strokes with temporal convolutional networks (TCN) \cite{kulkarni2021table}. Inspired from this work, this paper uses a similar approach where we design a regression network to deduce spin priors directly from human poses from a camera. 

\section{Factor Graphs For Asynchronous Ball Localization}

\begin{figure*}[ht!]
      \centering
      \includegraphics[width=0.8\linewidth]{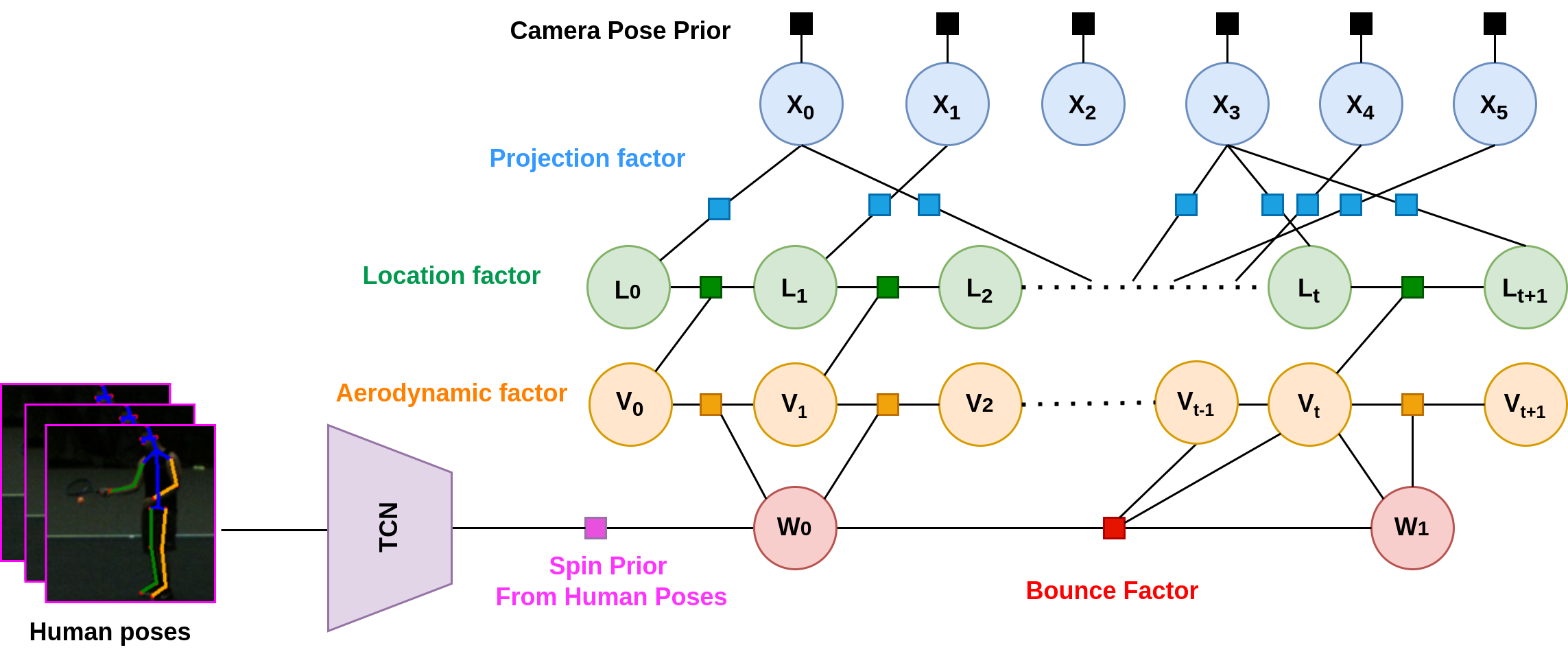}
      \caption{Example of the factor graph for ball localization at the $t+1$ time stamp. All the factors are colored squares and all the variables are labeled with circles, where $X_i$ is the $i^{th}$ camera pose, $L_t$ is the 3D location of the tennis ball at time step $t$, $V_t$ is the 3D velocity of the tennis ball at time step $t$, $W_j$ is the 3D spin of the tennis ball before the $j^{th}$ bounce. If detection is available in the queue, the factor graph will first expand }
      \label{fig: factor graph overview}
\end{figure*}

Factor graphs for asynchronous ball localization can be viewed as a graph where the size grows as the number of camera detections increases. Once the connections of the factor graph is updated, the location, velocity and spin of the ball can then be inferred. Since every single detection can update the factor graph and have a localization result from inference, this process is asynchronous without any pairing. 

\subsection{Variables and Factors}

The variables for our factor graph are shown in Figure \ref{fig: factor graph overview}. Variables are labeled with circles, where $X_i$ is the $i^{th}$ camera pose, $L_t$ is the 3D location of the tennis ball at time step $t$, $V_t$ is the 3D velocity of the tennis ball at time step $t$, $W_i$ is the 3D spin of the tennis ball before the $i^{th}$ bounce. 

The factors are colored with squares in the Figure \ref{fig: factor graph overview}. Factors can be thought of as functions that take the values of their associated variables as input and produce a scalar value or probability that quantifies the compatibility of those values with the factor's constraint (usually associated with an error function). In the proposed factor graph, the camera poses, $X$, have camera pose priors (black squares) which are computed from extrinsic calibration. The 3D ball locations, $L$, and the observing camera poses, $X$, have projection factor (blue squares). The adjacent location changes from previous time stamp $L_{t-1}$ to current time stamp $L_{t}$ is determined by the previous velocity, $V_{t-1}$, which form the location factor (green squares). The current velocity, $V_{t}$, is affected by previous velocity, $V_{t-1}$, as well as the previous spin, $W_0$, which is derived from aerodynamics. This factor is referred to as aerodynamic factor (orange square). For the spin, we assume the spin, $W$, of the ball only change after bounce from the ground. Therefore, if bouncing is detected, we will form a bounce factor (red square), connecting previous velocity, $V_{t-1}$, and spin, $W_0$, with current velocity, $V_{t}$, and spin, $W_1$. Finally, we also have our initial spin with a spin prior (pink square), which can be computed from human poses before hitting the ball. If human pose data are not available, we can set $W_0$ to zero with  large noise.

The whole factor graph can be factorized using (\ref{eq: graph factorization})

\begin{equation}
    G(\eta) = \prod_{i=0}^{M-1} f_i(\xi_i)
    \label{eq: graph factorization}
\end{equation}
where $\eta$ represent all variables in the graph, $f_i$ is the potential function of the $i^{th}$ factor, $\xi_i$ is all the variables connected to the $i^{th}$ factor, $M$ is the total number of factors in the graph. The potential function can be written as (\ref{eq: factor})
\begin{equation}
    f_i(\xi_i) = \exp(-\frac{1}{2}\|err_i(\xi_i)\|^2_{\Sigma_i})
    \label{eq: factor}
\end{equation}
where  $err_i$ is the error function for the $i^{th}$ factor, $\Sigma_i$ is the covariance matrix for the $i^{th}$ factor. Based on geometric or physics constraints, we derive our error function for each factor in the following sections. 

\subsubsection{Projection factor}
The 3D to 2D projection on camera can be written as
\begin{equation}
\pi^{proj}(L) = K\begin{bmatrix}
             R & T
         \end{bmatrix}
         L
\end{equation}
where $K$ is camera intrinsics, $R$ and $T$ are camera extrinsics (or camera pose $X$). Therefore, error function for projection factor can be written as
\begin{equation}
    err^{proj}(\xi) = u_{detect} - \pi(L)
\end{equation}
\subsubsection{Location factor}
The location factor can be view as a integration of velocity at small time interval. We use Euler formula to compute the next time stamp location, written as
\begin{equation}
    \pi^{loc}(L_s,V_s) = L_{s} + V_{s}\left[t(s+1) - t(s)\right]
\end{equation}
Therefore, error function for location factor is derived as
\begin{equation}
    err^{loc}(\xi) = L_{s+1} - \pi^{loc}(L_s,V_s)
\end{equation}
\subsubsection{Aerodynamic factor}
We use the formula derived in \cite{cooke2000overview} to compute the air drag and air lift \cite{cross2014measurements}, which can be written as 
\begin{equation}
    F_{D}(V) = -\frac{1}{2}\rho_{air}A C_D \|V\|V
\end{equation}

\begin{equation}
    F_{L}(V,W) = \frac{1}{2}\rho_{air} A C_L \|V\|^2 U
\end{equation}

where $U$ is the cross product $W \times V$, $C_D$ and $C_L$ are coefficients for air drag and air lift, respectively, A is the area of the ball, $\rho_{air}$ is the air density. Using the Newton's second law, we can the derive the acceleration with
\begin{equation}
    acc(V,W) = (F_D(V) + F_L(V,W))/m + g
    \label{eq:acc}
\end{equation}
\begin{equation}
    \pi^{aero}(V_s,W_j) = V_s + acc(V_s,W_j)\left[t(s+1) - t(s)\right]  
\end{equation}
Therefore, the error function for the velocity factor is
\begin{equation}
    err^{aero}(\xi) = V_{s+1} -\pi^{aero}(V_s,W_j)
\end{equation}
\subsubsection{Bounce factor}
For the bounce factor, we extend the formula derived in \cite{cross2002measurements} to estimate the outbound velocity and spin in 3D space as described in
\begin{equation}
    \pi^{bc}(\xi) =\begin{bmatrix}
        V_{x,t} & V_{y,t} & V_{x,t} & W_{x,j} & W_{y,j}  &W_{z,j} 
    \end{bmatrix}^T\\
    \label{eq: bounce}    
\end{equation}
where 
\begin{eqnarray}
    V_{x,t} &=& \frac{R W_{y,j-1} + (1.5 R ^2/R_1^2)V_{x,t-1}}{1+1.5 R ^2/R_1^2 }\\
    V_{y,t} &=& \frac{-R W_{x,j-1} + (1.5 R ^2/R_1^2)V_{y,t-1}}{1+1.5 R ^2/R_1^2 }\\
    V_{z,t} &=& - e_{z} V_{z,t-1}\\
    W_{x,j} &=& -V_{y,t} / R\\
    W_{y,j} &=& V_{x,t} / R\\
    W_{z,j} &=&  W_{z,j-1} 
\end{eqnarray}
where $R$ is the outer radius of tennis ball and $R_1$ is the average radius of  the shell, $e_z$ is the restitution of the tennis ball. Therefore, the error function for the bounce factor is
\begin{equation}
    err^{bc}(\xi) = [V_{s+1},W_{j+1}]^T -\pi^{bc}(V_s,W_j)
\end{equation}

\subsection{Incremental Inferences}
As illustrated in Figure \ref{fig: factor graph overview}, the depth of the factor graph is growing linearly with the number of camera detections. Each time a new detection is completed and transmitted to the central computer, we will first add new connections to the factor graph, then update values for all the variables in the graph. This approach avoids the need for a synchronization process or pairing, indicating that irrespective of the camera responsible for the detection and thereby achieving a adaptable system. 

Factor graph inferences need to solve the maximize likelihood optimization problem.
Since the size of the factor graph will become very large as more detections come, which makes traditional optimizers impossible to run in real time. In this paper, we use the factor graph package, GTSAM, to solve the factor graph we build as in Figure \ref{fig: factor graph overview}. This package use ISAM2 algorithm \cite{kaess2008isam,kaess2012isam2}, which can relinearize and reorder the factor graph so that the graph can be optimized incrementally, and hence achieve real time inference speed. The linearization process is described in equation (\ref{eq: factor graph linearization})
\begin{equation}
     \operatorname*{argmin}_\eta -\log(G(\eta)) \approx \operatorname*{argmin}_\eta \|J\eta - b\|^2
     \label{eq: factor graph linearization}
\end{equation}
where the Jacobian $J$ is the error function with respect to the associated factor's variables. 

\subsection{Prediction}
The prediction of the ball will use the dynamic equations we derived in (\ref{eq:acc}) and use Runge-Kutta $4^{th}$ order to integrate. The ODE for ball during flight is written as
\begin{equation}  
    \begin{bmatrix}
        \dot{L}\\\dot{V}\\ \dot{W}
    \end{bmatrix}
    = \begin{bmatrix}
        V \\ acc(V,W)\\0
    \end{bmatrix}
\end{equation}

The integration will also check if the ball bounce on the ground ($V_z <0 \text{ and } Lz < z_{\text{ground}}$). If the bounce is detected, we will terminate the current ODE integration and apply the bounce dynamics to the current state velocity and spin as derived in equation (\ref{eq: bounce}), followed by a new integration  starting with the new velocity and spin.

\section{Spin Priors from Human Poses}

\begin{figure}[t]
      \centering
      \includegraphics[width=0.94\linewidth]{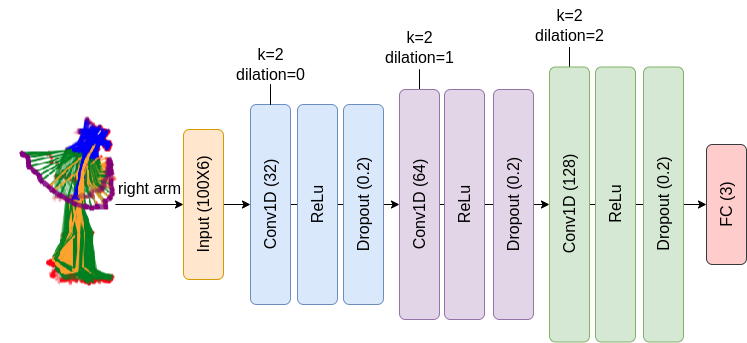}
      \caption{TCN for spin prior regression with human pose sequences. The model takes time-series data of human poses when hitting the ball as input and outputs an estimated spin value for the ball.}
      \label{fig: TCN}
    \end{figure}
    
In this section, we outline our method for deriving spin priors from human poses using a neural network.


\subsection{Data Preparation}

To provide input for our neural network, we employ AlphaPose \cite{alphapose} to detect human poses in images. Each trajectory is linked with 100 frames of human poses, emphasizing the key points on the right arm: shoulder, elbow, and wrist. These poses are centered at the ``hip.''
 

We introduce an iterative approach to label the spins in each trajectory. Factor graph inference, known for its capacity to update variable values within the graph, gains confidence with accumulating detections. However, initially, our factor graph lacks knowledge about suitable spin priors. To tackle this, we start with initial spin estimations and conduct inference across all detections until the trajectory's end. As the initial spin estimations evolve, we utilize the inference from the final detection as input for a new iteration. This iterative process continues until the spin estimate converges to a value that is no longer changing significantly, which is then used to label the spin for each trajectory.


\subsection{Neural Network Model}

For sequence modeling, we use temporal convolutional neural networks (TCNs) to estimate spin priors from sequential human poses using regression. Figure~\ref{fig: TCN} illustrates the architecture of our model. We empirically found that TCNs outperformed recurrent neural networks (RNNs) and long short-term memory networks (LSTMs) for this task, which is supported by prior work \cite{bai2018empirical}.

    
\section{Experiments and Results}
\subsection{Hardware Setups}

 \begin{figure}[t]
      \centering
      \includegraphics[width=0.9\linewidth]{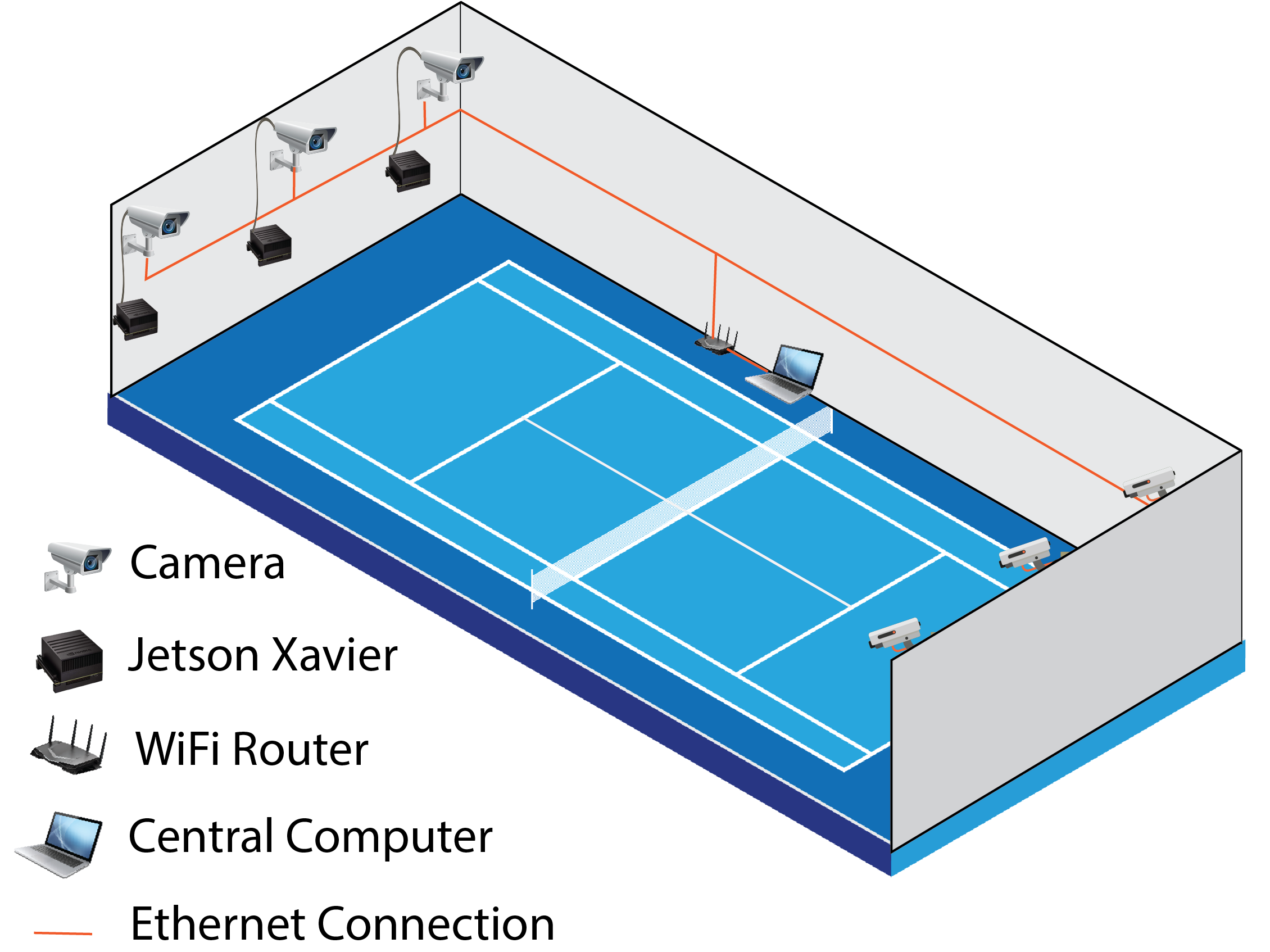}
      \caption{Camera setup on an indoor tennis court.}
      \label{fig:camera_setup}
    \end{figure}

In order to track and detect the tennis ball on an indoor tennis court, we use a multi-cameras setup (Figure \ref{fig:camera_setup}). The system consists of six Blackfly S cameras, each connected to a Jetson Xavier. The cameras are mounted on the back walls of the court at a height of 5.5 meters, looking down at the court. Three cameras are mounted on each side of the court, providing overlapping coverage of the entire court. This setup maximizes the chances of the ball being detected by multiple cameras, and allows us to track the ball from various angles, which is more robust than having multiple cameras look at the ball from the same angle~\cite{optitrack_camera_placement}. The intrinsic calibration of each camera is performed using a checkerboard. For extrinsic calibration, an AprilTag~\cite{apriltag} is placed on the court at a known position. Using the AprilTag detection and the detection of keypoints on the tennis court, the extrinsic calibration of each camera is performed.

The Jetson Xaviers acquire color images from the cameras. The detection algorithm then performs background subtraction and noise removal to isolate the moving parts of the image. This allows us to focus on the moving ball and removes the need to process the parts of the image not containing the moving ball. We then run a YOLO tiny~\cite{huang2018yolo} model trained to detect tennis balls on the isolated parts of the image. The image acquisition and ball detection algorithm runs at $\sim$140 Hz on each Jetson Xavier.

We use Ethernet communication to share ball detections from the Jetson Xaviers to the router and central computer, which uses the detections to run our ball trajectory estimation algorithm. This results in an average latency of 1 millisecond between the Jetson Xaviers sending the detections and the central computer receiving them. This represents a substantial improvement compared to a prior decentralized multi-camera configuration that relied on WiFi-based communication, as outlined in the work by Zaidi et al.~\cite{zaidi2023athletic}. In their setup, they achieved a detection rate of 25 Hz per camera and exhibited a considerably higher average latency of 110 milliseconds between the Jetson Nano and the central computer.

\subsection{Spin priors from Human Poses}

\begin{figure}[t]
      \centering
      \includegraphics[width=0.79\linewidth]{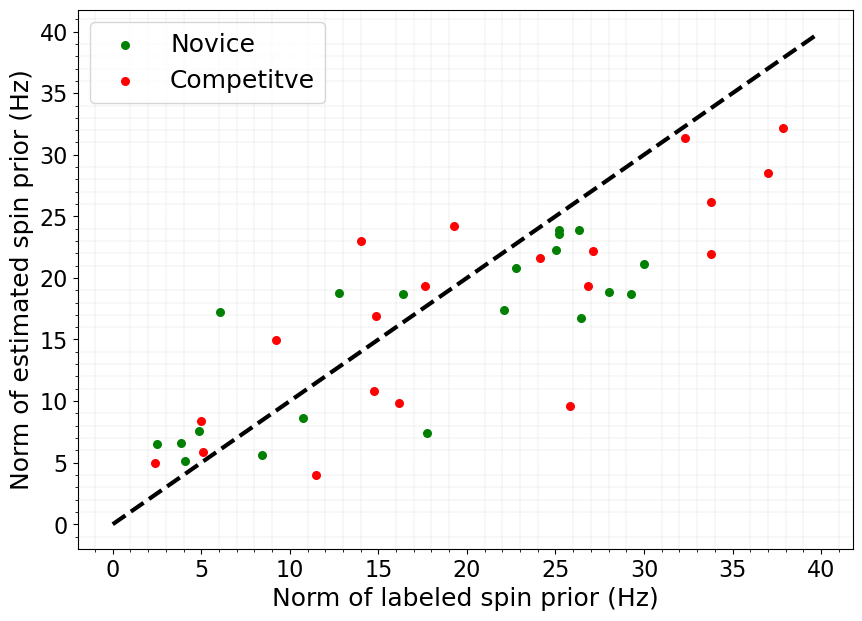}
      \caption{Scatterplot of labeled vs. estimated spin priors in the validation dataset, showing a strong correlation.}
      \label{fig: spin estimation scatter plot}
    \end{figure}

We partitioned our dataset into training and validation sets at an 80:20 ratio, aiming to provide a robust training ground for our model while retaining a substantial portion for validation. The Temporal Convolutional Network (TCN) was utilized as the backbone of our neural network, chosen for its prowess in handling sequential data effectively. To quantify the performance of our model, we utilized the Root Mean Square Error (RMSE) metric. This choice is grounded in its efficacy in measuring the average magnitude of error between predicted and actual values, which is particularly pertinent in our context of estimating spin priors. Our model demonstrated promising results with a training RMSE of 2.95 Hz and a validation RMSE of 5.27 Hz. Figure \ref{fig: spin estimation scatter plot} shows the scatter plot of the spin prior estimation across the validation dataset. The figure illustrates a linear distribution between the labeled spin prior and the estimated spin prior. Notably, both novice and competitive players exhibit strong linear relationships, with Pearson correlation coefficients ($R$) of 0.807 and 0.815, respectively, and highly significant $p < 0.001$ values for both scenarios. 
   
\subsection{Prediction Results}
The primary objective of the trajectory prediction  is to facilitate advanced motion planning for the tennis robot, thereby enhancing the likelihood of successfully returning a tennis ball. In this section, we focus on assessing the accuracy of predictions pertaining to both the initial and the second landing positions of the ball. These two landing positions are chosen because the wheelchair tennis rules permit players to strike the ball either before its initial landing or prior to the second landing. Thus, they hold significant relevance and are representative metrics in the domain of wheelchair tennis gameplay. 

To assess the performance of our approach, we conduct several comparative evaluations. Firstly, we compare it with the method proposed in \cite{zaidi2023athletic}, which employed a similar multi-camera decentralized setup and utilized an Extended Kalman Filter (EKF) for tennis ball trajectory estimation and prediction. We also explore an alternative by using an adaptive EKF (AEKF) that models the spin as a hidden state instead of using the standard EKF. Furthermore, we conduct an ablation study by removing the spin priors from our approach. Subsequently, we compare our method's performance when employing the estimated spin from the TCN against using the spin labels directly.

\begin{figure}[t]
      \centering
      \includegraphics[width=0.99\linewidth]{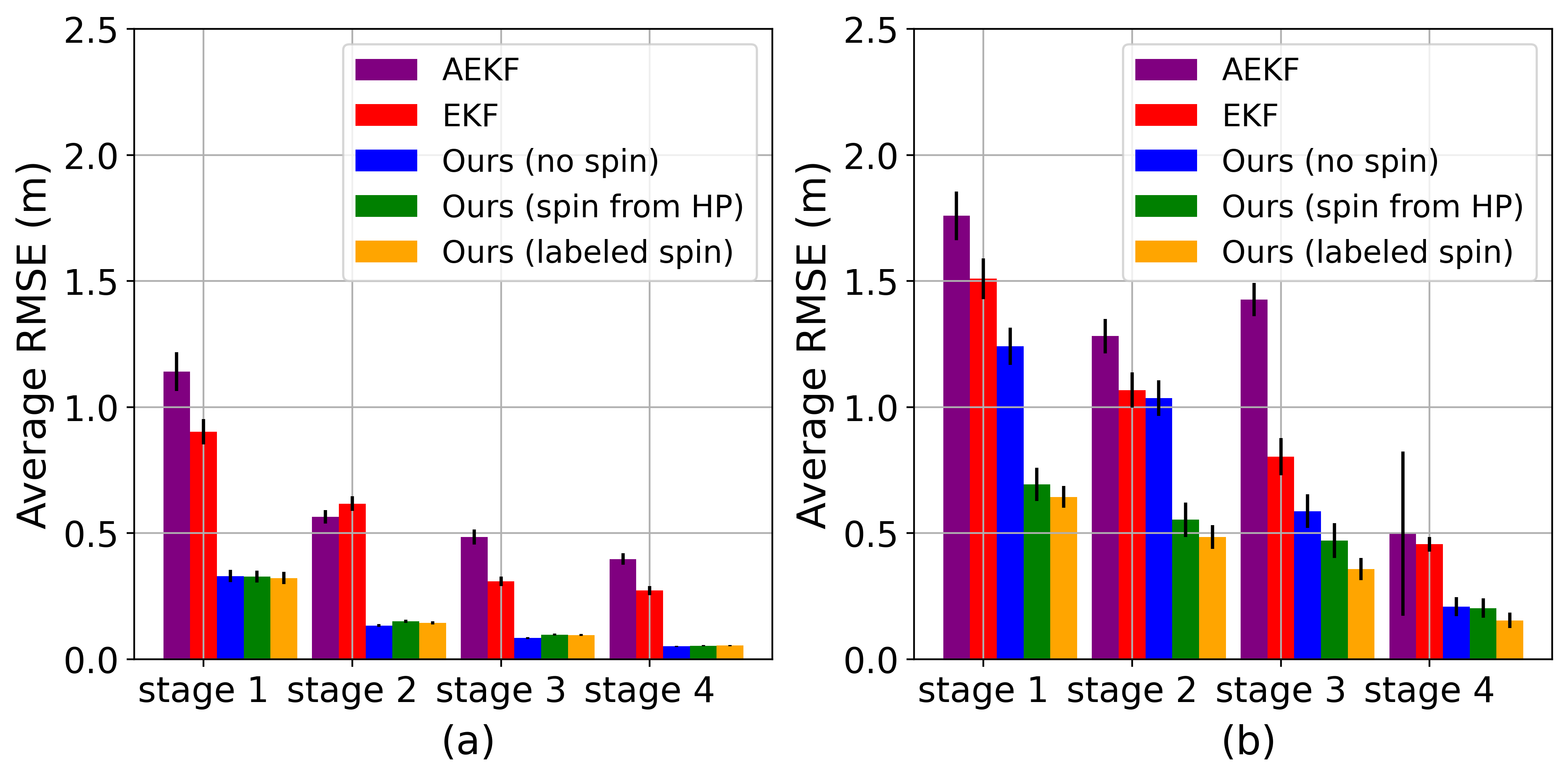}
      \caption{The average RMSE of 4 stages for the prediction of (a) the first landing position and (b) the second landing position.}
      \label{fig:4_stage_landing_rmse}
      \vspace{-4mm}
    \end{figure}

In order to analyze how prediction error changes with different number of observations from camera, we bin each ball trajectory evenly into quartiles or ``stages:'' Stage 1 contains the first 25\% of the observations; Stage 2 contains the next 25\% of the observations, etc. We then analyze the mean and standard deviation of the RMSE for each stage (Figure \ref{fig:4_stage_landing_rmse}). For example, Stage 1 results average across the accuracy of ball estimation from when there is only a single detection until a total of 25\% of detections for that trajectory is available to the algorithm. We chose this binning procedure as ball detections happen at a non-uniform cadence (i.e., asynchronously) and the trajectories are heterogeneous in duration and in the timing of the bounces.

The results for prediction error shown in Figure \ref{fig:4_stage_landing_rmse} show that the prediction error reduces across time, with the biggest drop occurring after the first 25\% of the trajectory has been completed (from Stage 1 to Stage 2). Furthermore, it's notable that incorporating our TCN for spin prediction from human pose data (Figure \ref{fig:4_stage_landing_rmse}, green) significantly enhances the accuracy of second bounce location prediction (Figure \ref{fig:4_stage_landing_rmse}(b)) compared to the baseline approaches, AEKF (purple) and EKF (red). It also outperforms our approach without considering spin (blue) and closely approaches the trajectory prediction achieved using the labeled spin values (orange).
The advantage of incorporating spin information becomes evident primarily in predicting the second bounce, as the factor graph can effectively handle in-flight spin dynamics but requires spin details to accurately model the impact of spin on the ball's first bounce. This hypothesis gains support from the observation that the no-spin model (blue) shows a significant improvement in predicting the second bounce location (Figure \ref{fig:4_stage_landing_rmse}(b)) in Stages 3-4, which typically occur after the initial bounce.

\begin{figure}[t]
      \centering
      \includegraphics[width=0.99\linewidth]{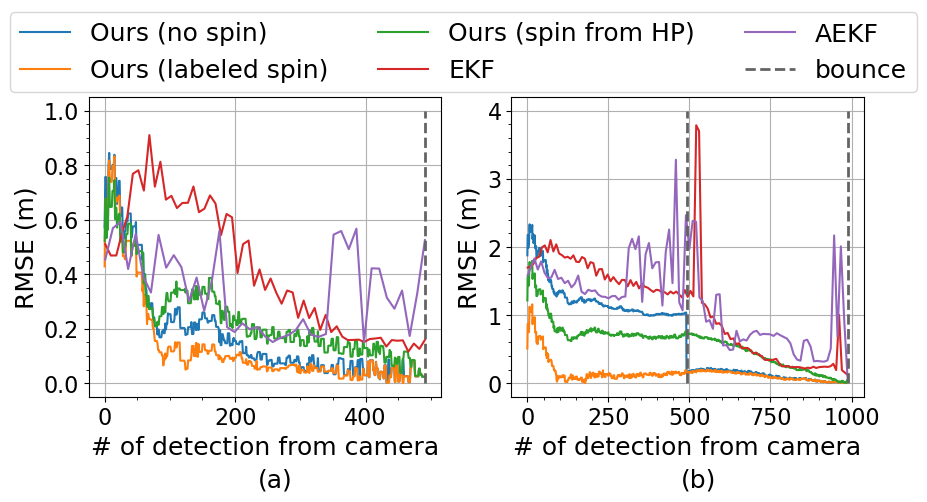}
      \caption{RMSE results with respect to number of observation (a) initial landing position prediction errors (b) second landing position errors for a single shot with heavy spin played by a professional tennis player.}
      \label{fig:landing_errors}
\end{figure}

To delve deeper into the potential impact of Magnus and lift forces resulting from spin, especially prevalent among highly skilled tennis players renowned for employing powerful spins, we enlisted the expertise of a competitive tennis player to hit shots with substantial spin. Assessment of the performance of our approach (green) in accurately predicting the locations of the first and second bounces is illustrated in Figure \ref{fig:landing_errors}, where we also compare it with baseline methods for a single trajectory. The most accurate predictions are achieved when our approach is combined with labeled spin values (orange). However, our neural network model faced challenges in accurately approximating spin priors, likely due to limited training data and the absence of information regarding the racket's pose. Nonetheless, our approach consistently outperformed the baseline EKF (red) and AEKF (purple) methods.

It is important to highlight that incorporating information about the racket's pose could enhance spin estimation accuracy, complementing insights derived from the analysis of human poses. Overall, the substantial improvement achieved through the use of labeled spin priors underscores the immense potential of accurate spin estimation. 

\section{LIMITATIONS AND FUTURE WORK}

We note key limitations in our approach for ball trajectory prediction. When dealing with balls that have been struck with a substantial spin, our model encounters challenges in precisely estimating the spin prior from the human pose data. This limitation stems from the small amount of training data and the absence of racket pose details and the type of grip used (e.g., Eastern, Western, or Continental). Despite these challenges, the prediction using labeled spin priors elucidate a substantial potential to enhance the accuracy of landing position predictions if spin parameters can be precisely extrapolated from human postures, indicating promising avenues for further improvement. Further, we found that we had to perform ad hoc modifications for the coefficients associated with the bounce dynamics equations from prior work to be robust to high-spin dynamics (e.g., ball-ground slip). In future work, we propose to augment our factor graphs to have a learnable factor for bouncing.

\section{CONCLUSIONS}
   In this study, we introduce an innovative asynchronous approach to predicting the state of a tennis ball utilizing a factor graph. To facilitate precise inferences at the initial stages and predict the landing positions, we have also developed a neural network utilizing Temporal Convolutional Network (TCN) to infer spin priors from human postures. Our findings demonstrate that the algorithm can proficiently and accurately estimate the ball's location, yielding smooth trajectory plots.

\section*{ACKNOWLEDGMENT}
This work was supported by a grant from Google. Special thanks to Dr. Pannag Sanketi and Dr. David D'Ambrosio for their invaluable support and insightful feedback throughout the duration of this work.



\bibliographystyle{IEEEtran}
\bibliography{References}

\end{document}